\documentclass{article}
\usepackage{spconf,amsmath, graphicx}
\usepackage{threeparttable}
\usepackage{graphics}
\usepackage{stfloats}
\usepackage[linesnumbered,boxed,ruled,commentsnumbered]{algorithm2e}
\usepackage{float}
\usepackage{color}
\usepackage{hyperref}
\usepackage{appendix} 
\hypersetup{
            colorlinks=true,
            linkcolor=red,
            anchorcolor=blue,
            citecolor=red}

\definecolor{zmblue} {rgb}{0.19, 0.55, 0.91}

\title{Anomaly Detection via Self-organizing Map}
\name{Ning Li$^{1,\star}$, Kaitao Jiang$^{2,\star}$\thanks{$\star$ \ Ning Li and Kaitao Jiang are co-first authors.}, Zhiheng Ma$^{2}$, Xing Wei$^{1,\S}$\thanks{$\S$ \ Corresponding author: Xing Wei.}, Xiaopeng Hong$^{2}$, Yihong Gong$^{1, \dagger}$\thanks{$\dagger$ \ IEEE Fellow}
}
\address{ $^1$School of Software Engineering, Xi'an Jiaotong University \\
        $^2$Faculty of Electronic and Information Engineering, Xi'an Jiaotong University \\
        \small Email: {\{ddzxlining, jkt1143958845, mazhiheng\}@stu.xjtu.edu.cn, \{weixing, hongxiaopeng, ygong\}@mail.xjtu.edu.cn}
        }

\begin{document}
%\ninept
%
\maketitle
\begin{abstract}
Anomaly detection plays a key role in industrial manufacturing for product quality control.
Traditional methods for anomaly detection are rule-based with limited generalization ability. 
Recent methods based on supervised deep learning are more powerful but require large-scale annotated datasets for training. 
In practice, abnormal products are rare thus it is very difficult to train a deep model in a fully supervised way.
In this paper, we propose a novel unsupervised anomaly detection approach based on Self-organizing Map (SOM). Our method, Self-organizing Map for Anomaly Detection (SOMAD) maintains normal characteristics by using topological memory based on multi-scale features. SOMAD achieves state-of-the-art performance on unsupervised anomaly detection and localization on the MVTec dataset.
\end{abstract}
\begin{keywords}
anomaly detection, self-organizing map, anomaly localization
\end{keywords}
\section{Introduction}
\label{sec:introduction}

Anomaly detection is a classical problem and refers to identifying those samples in dataset that are significantly different from normal samples.
Usually, the abnormal samples are not accessible at the training phase, or they are not sufficient enough to model its distribution, compared to its huge diversity. Moreover, there are some unexpected anomalies in the process of inference. Therefore, anomaly detection is still a challenging task.
Traditional machine learning methods demand to design different pipelines for each type of product, thus the generalization performance is unsatisfactory. Supervised deep learning methods require a large number of labeled samples, but in practice it is often difficult to satisfy. On the one hand, defective images are very rare in manufacturing lines. On the other hand, we cannot know all types of defects before they occur. On the contrary, the normal samples are almost the same and easily accessible. Therefore, modeling the normality of normal samples for abnormal detection using an unsupervised manner is a feasible idea. 
In recent years, a large number of anomaly detection methods have been proposed.
We provide an overview of existing unsupervised anomaly detection methods in the following. In order to better analyze this problem, we categorize these methods into either learning from scratch or leveraging pre-trained convolutional neural network (CNN).

\textbf{Learning from Scratch:}
Learning representations from scratch is a major way for image classification and other tasks. Autoencoder-based methods are the most widely used out of a large number of learning representation approaches in unsupervised anomaly detection.
Autoencoders (AE) \cite{MvtecAD,Student,MemoryAE,AE2,AE3,Mirrored_AE,AE4,AE5}, variational autoencoders (VAE) \cite{VAE1,VAE2,VAE3} or generative adversarial networks (GAN) \cite{Ganomaly,GAN1,fAnoGAN,superpixel_inpainting} try to learn the distributions of normal samples by training models to reconstruct normal images exclusively using normal images. These methods usually first compress input image to a low dimensional latent vector, and then the normal image is expected to be reconstructed by the latent vector. After that, anomalies can be detected by comparing the input image and its reconstruction at the pixel level.  
However, in the complex real-world datasets, autoencoder-based methods can yield good reconstruction results for abnormal images too \cite{Perera2019OCGANON}, which results in failing to distinguish between normal and abnormal images.

\textbf{Leveraging Pre-trained CNN:}
Usually the sample scale of anomaly detection datasets is relatively smaller than ImageNet dataset \cite{ImageNet}, so it is inadequate to learning a good representation from scratch. 
Many methods have also been proposed for anomaly detection \cite{Ma_distance,Bergman2020DeepNN,AggregatingDetection,reiss2020panda} or segmentation \cite{Student, PatchSVDD, Padim,ModelingPercep,rudolph2020differnet} by using deep representations pre-trained on ImageNet \cite{ImageNet}. They compare patch features at the same position from target and normal image to obtain pixel-level anomaly map. The image-level anomaly score will be calculated by aggregating the anomaly score of pixels in the image. 
However, these methods are not robust to unaligned datasets, especially for large scales of translation and rotation. 
Cohen et al. \cite{SPADE} proposed a method to tackle the problem of unaligned images, they search the most similar patch from all positions of the 50 nearest training images. However, it is computationally intensive because of the procedure of retrieval from a large normal feature gallery. 

To alleviate the aforementioned problems, we proposed a novel and efficient approach: self-organizing map for anomaly detection (SOMAD). It makes use of pre-trained CNN to extract the features of patches and leveraging the SOM to maintain the neighborhood relationship of embedding vectors in topology space. In addition, we have greatly reduced the search space by mapping the normal feature space into 2-dimensional space through SOM. Each patch position corresponds to a SOM node, and every patch in the training images is assigned to the nearest SOM node in topology space. 

Our major contribution can be summarized as follows:
\begin{itemize}
\item We proposed an effective and efficient method based on SOM for anomaly localization and detection. 
\item Our approach outperforms state-of-the-art methods for anomaly localization and detection on the MVTec AD dataset, especially on unaligned categories.
\end{itemize}

\section{Proposed Method}
\label{sec:method}
\subsection{Feature Extraction}
\label{Feature extraction}
Recently, Bergman et al. \cite{Bergman2020DeepNN} showed that the features learned using generic ImageNet-based feature extractors are competitive with self-supervised methods. Therefore, we extract feature through a ResNet pre-trained on the ImageNet dataset \cite{ImageNet}. 
Taking into account both global and local features, we combined different level features similarly to \cite{SPADE}. 
For a target image $x$, $f$ is the extracted feature maps from a pre-trained Wide-RestNet50-2 \cite{WideResNet}.
We divided an input image into a grid of $W\times H$, where $W$ and $H$ are the width and height of the largest feature map used to extract embeddings respectively. Finally, each patch $x_i$ at $i (i\in [1,W\times H])$ position in this grid has a one-to-one correspondence with feature embedding $f_{x_i}$.
\subsection{Memorizing Normality via Self-organizing Map}
\label{som}
\begin{figure}[!b]
	\centering
	\includegraphics[scale=0.4]{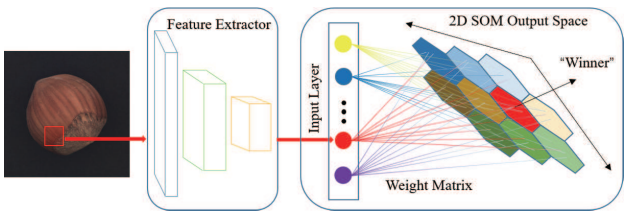}
	\caption{Illustration of memorizing normality via SOM.}
	\label{fig:som}
\end{figure}
Self-organizing Map (SOM) is one of the most popular artificial neural network (ANN) models, developed by \cite{SOM}. The main aim of SOM is to preserve the topology of high-dimensional data when they are mapped into lower dimensional space (usually two-dimensional).
It consists of an input layer, which is fed with the high-dimensional feature vector $f$, and a competitive layer, which contains $K\times K$ nodes organized in a regular 2D grid $M=\{m_k\}_{k=1}^{K^2}$, where $m_k$ denotes the centroid vector of node $k$. 
SOM maintains the neighborhood relationship of embedding vectors in topology space: similar embeddings are located closer while less similar ones gradually farther away. 
Given an input vector $f$, SOM finds the 'winner' node $c$ (also called best match unit, BMU) by picking the centroid vector $m_c$ closest to $f$:
\begin{equation}
	 c=\mathop{\arg\min}\limits_{k} \lVert f-m_k \rVert ^2_2,
	\label{eq:som winner}
\end{equation}
During the training phase, all patch features of entire defect-free training images are extracted and formed a feature gallery $G$. 
After that, BMU of each input is found, the next step is to update weights matrix. It should be noted that we not only update the weight of BMU, but also its neighbors'.
After training convergence, each patch is assigned to one centroid, which can represent the set of the feature vectors closest to it. Fig. \ref{fig:som} shows the process of memorizing normality via SOM.

\begin{algorithm}[!t]
    \SetAlgoNoLine
    \SetKwInOut{Input}{Input}
    \SetKwInOut{Output}{Output}
    \Input{
    Training feature gallery $G$, SOM map size $K$, test images $X$
    }
    \Output{
    SOM weight matrix $M$, anomaly map
    }
    \BlankLine
    \tcp{Training stage}
    Initialize an SOM of $K \times K$ nodes  $M$ \;
    Train $M$ on $G$ as discribed in Section \ref{som}  \;
    \For{$k=1$ to $K^2$}{
        Compute $\sigma_i$ using Equation. \eqref{eq:sigma}  ;\
    }
    \tcp{Inference stage}
    \For{test image $x \in X$}{
        Extract the feature $f_x$ of $x$ \;
        \For{patch feature $f_{x_i} \in f_x$}{
            Retrieve the top $k$ nearest SOM centroid nodes of $f_{x_i}$ \;
            Compute anomaly score using Equation. \eqref{eq:ma_distance} \;
        }
        Compute the anomaly score of the image $x$ as described in Section \ref{anomaly_map} \
    }
    \caption{The pipeline of SOMAD}
    \label{alg:pipline}
\end{algorithm}

\subsection{Anomaly Map Computation}
\label{anomaly_map}
After memorizing all normal patch features on the training set through SOM, we can calculate the anomaly score for each patch $x_i$. 
Firstly, for each SOM centroid node $m_i$ in $M$, we collect all the normal patch features matched with $m_i$ to form a set $G_i$. After that, we calculate the covariance $\sigma_i$ of $G_i$ as follows:
\begin{equation}
    \label{eq:sigma}
    \sigma_i = \frac{1}{N-1}\sum_{j=1}^N(f_{x_i}^j-m_{i})(f_{x_i}^j-m_{i})^T+\epsilon I,
\end{equation}
where $N$ is the size of $G_i$. To ensure $\sigma_i$ full rank and invertible, a regularisation term $\epsilon I$ is added. 
For a test image $x \in X$, we first extract the features $f_{x_i}$ of each patch as described in Section. \ref{Feature extraction}. 
Then, the top $k$ nearest SOM centroid nodes of patch $x_i$: $N_k{(x_i)}$ are picked according to Equation \eqref{eq:som winner}. Inspired by PaDiM \cite{Padim}, we also use the Mahalanobis distance \cite{Mahalanobis} $M_i$ to calculate the anomaly score of patch $x_i$, where $M_i$ is computed as follows:
\begin{equation}
    M_i = \sqrt{(f_{x_i}-m_{i})^T\sigma_i^{-1}(f_{x_i}-m_{i}}.
    \label{eq:ma_distance}
\end{equation}
Finaly, the anomaly score $s_i$ of patch $x_i$ is the minimum Mahalanobis distance between the feature of $x_i$ and $N_k{(x_i)}$ . 
Further, we can obtain image-level anomaly score by calculating the maximum score among abnormal pixels. The details of these specific parameters mentioned above will be described in Section. \ref{detail}. Algorithm. \ref{alg:pipline} shows the process of SOMAD.

\section{Experiments}
\label{sec:experiments}

\subsection{Datasets and Evaluation Metrics}
\label{dataset_metrics}
We evaluate our method following the standard protocol where no defective images are used in training on MVTec AD dataset.

\textbf{Datasets.}
MVTec AD \cite{MvtecAD} is a real-world dataset for unsupervised industrial anomaly detection and localization. It consists of 15 classes (5 for texture and 10 for object) industrial images, and the image resolution ranges from $700\times700$ to $1024\times1024$. There are 3629 defect-free images for training and 1725 images for testing. 
The test set contains both defective images and defect-free images. Each class has one or more types of defects labeled at the segmentation level. 

\textbf{Metric.}
We compute two threshold independent metrics to assess the localization performance. One of them is pixel-level Area Under the Receiver Operating Characteristic curve (AUROC) score, where the true positive rate is the percentage of pixels correctly classified as anomalous. Another metric is the per-region-overlap score (PRO-score) \cite{Student}. It consists in plotting a curve of mean values of the correctly classified pixel rates as a function of the false positive rate between 0 and 0.3 for each connected component. The PRO-score is the normalized integral of the curve. A high PRO-score means that anomalies of different scales are well-localized with few false positives. We also give the image level AUROC, where the true positive rate is the percentage of images correctly classified as anomalous to assess the detection performance.

\subsection{Implementation Details}
\label{detail}

We use a Wide-Resnet50x2 feature extractor, which is pre-trained on ImageNet datasets. We resized the images to $256\times256$ and center crop them to $224\times224$. 
Similar to \cite{SPADE}, we extracted feature maps of the first three layers, and the dimensions are 256, 512, 1024, respectively. 
In order to better encode the patch, we combine information from different semantic levels. Specifically, the features of the last two layers are upsampled to 56x56, and then the three layers feature maps were concatenated together in the dimension to form the patch embeddings. Each patch corresponds to a 1792 dimensions embedding code. We use SOM of size $56\times56$ for all datasets. We initialize the nodes of SOM using the average of patches at the same position from different training images.
A Gaussian filter on the anomaly maps with parameter $\sigma=4$ like in \cite{SPADE} was used to smooth the results. 
In all experiments we used $k=4$. 

\subsection{Experimental Results and Discussions}

\label{result}
\def\arraystretch{0.9}
\begin{table}[!t]
    \footnotesize
	\centering
	\caption{Pixel level anomaly localization accuracy on MVTec (AUROC\%).}
	\label{pixel_auc_mvtec}
	\setlength{\tabcolsep}{1mm}{
	\begin{tabular}{c|ccccc|c}
		\hline
		Model & AE SSIM & SMAI & Patch\cite{PatchSVDD} & SPADE & PaDiM &  SOMAD\\
		& \cite{MvtecAD},\cite{Student} & \cite{superpixel_inpainting} & SVDD & \cite{SPADE} & \cite{Padim} & \textbf{Ours} \\
		\hline
		Bottle     & 93.0 & 86.0 & 98.1 & 98.4 & 98.3 & 98.3 \\
		Cable      & 82.0 & 92.0 & 96.8 & 97.2 & 96.7 & 98.2 \\
		Capsule    & 94.0 & 93.0 & 95.8 & 99.0 & 98.5 & 98.7 \\
		Carpet     & 87.0 & 88.0 & 92.6 & 97.5 & 99.1 & 98.9 \\
		Grid       & 94.0 & 97.0 & 96.2 & 93.7 & 97.3 & 98.4 \\
		Hazelnut   & 97.0 & 97.0 & 97.5 & 99.1 & 98.2 & 98.4 \\
		Leather    & 78.0 & 86.0 & 97.4 & 97.6 & 99.2 & 99.1 \\
		Metal\_nut & 89.0 & 92.0 & 98.0 & 98.1 & 97.2 & 98.0 \\
		Pill       & 91.0 & 92.0 & 95.1 & 96.5 & 95.7 & 98.0 \\
		Screw      & 96.0 & 96.0 & 95.7 & 98.9 & 98.5 & 99.1 \\
		Tile       & 59.0 & 62.0 & 91.4 & 87.4 & 94.1 & 94.8 \\
		Toothbrush & 92.0 & 96.0 & 98.1 & 97.9 & 98.8 & 98.5 \\
		Transistor & 90.0 & 85.0 & 97.0 & 94.1 & 97.5 & 95.3 \\
		Wood       & 73.0 & 80.0 & 90.8 & 88.5 & 94.9 & 94.4 \\
		Zipper     & 88.0 & 90.0 & 95.1 & 96.5 & 98.5 & 98.7 \\
		\hline
		Average    & 87.0 & 89.0 & 95.7 & 96.5 & 97.5 & \textbf{97.8} \\
		\hline
	\end{tabular}}
\end{table}
\begin{table}[!h]
\footnotesize
	\centering
	\caption{Pixel level anomaly localization accuracy on MVTec (PRO-score\%).}
	\label{pixel_pro_mvtec}
	\setlength{\tabcolsep}{1mm}{
	\begin{tabular}{c|cccc|c}
		\hline
		Model & AE SSIM & VAE & SPADE & PaDiM &  SOMAD \\
		& \cite{MvtecAD},\cite{Student} & \cite{Padim} & \cite{SPADE} & \cite{Padim} & \textbf{Ours} \\
		\hline
		Bottle     & 83.4 & 70.5 & 95.5 & 94.8 & 94.7 \\
		Cable      & 47.8 & 77.9 & 90.9 & 88.8 & 93.4 \\
		Capsule    & 86.0 & 77.9 & 93.7 & 93.5 & 93.4 \\
		Carpet     & 64.7 & 61.9 & 94.7 & 96.2 & 95.5 \\
		Grid       & 84.9 & 40.8 & 86.7 & 94.6 & 95.3 \\
		Hazelnut   & 91.6 & 77.0 & 95.4 & 92.4 & 95.1 \\
		Leather    & 56.1 & 64.9 & 97.2 & 97.8 & 97.7 \\
		Metal\_nut & 60.3 & 57.6 & 94.4 & 85.6 & 93.6 \\
		Pill       & 83.0 & 79.3 & 94.6 & 92.7 & 96.5 \\
		Screw      & 88.7 & 66.4 & 96.0 & 94.4 & 96.0 \\
		Tile       & 17.5 & 24.2 & 75.9 & 86.0 & 81.3 \\
		Toothbrush & 78.4 & 85.4 & 93.5 & 93.1 & 90.7 \\
		Transistor & 72.5 & 61.0 & 87.4 & 84.5 & 91.6 \\
		Wood       & 60.5 & 57.8 & 87.4 & 91.1 & 88.2 \\
		Zipper     & 66.5 & 60.8 & 92.6 & 95.9 & 95.9 \\
		\hline
		Average    & 69.4 & 64.2 & 91.7 & 92.1 & \textbf{93.3} \\
		\hline
	\end{tabular}}
\end{table}
\begin{table}[htb]
\footnotesize
	\centering
	\caption{Image level anomaly detection accuracy on MVTec (AUROC\%).}
	\label{img_auc}
	\begin{tabular}{c|ccc|c}
	    \hline
		Model   & Patch\cite{PatchSVDD} & SPADE & PaDiM & SOMAD \\
		 &  SVDD  & \cite{SPADE} & \cite{Padim} & \textbf{Ours} \\
		\hline
		Bottle      & 98.6  & 97.2 & 99.8 & \textbf{100}  \\
		Cable       & 90.3  & 84.4 & 92.2 & \textbf{98.8} \\
		Capsule     & 76.7  & 89.7 & 91.5 & \textbf{93.8} \\
		Carpet      & 92.9  & 92.8 & 99.9 & \textbf{100}  \\
		Grid        & 94.6  & 47.3 & 95.7 & 93.9 \\
		Hazelnut    & 92.0  & 88.1 & 93.3 & \textbf{100}  \\
		Leather     & 90.9  & 95.4 & 100 &  \textbf{100}  \\
		Metal\_nut  & 94.0  & 71.0 & 99.2 & \textbf{99.7} \\
		Pill        & 86.1  & 80.1 & 94.4 & \textbf{98.6} \\
		Screw       & 81.3  & 66.7 & 84.4 & \textbf{95.5} \\
		Tile        & 97.8  & 96.5 & 97.4 & \textbf{98.7} \\
		Toothbrush  & 100.0 & 88.9 & 97.2 & 98.6 \\
		Transistor  & 91.5  & 90.3 & 97.8 & 94.5 \\
		Wood        & 96.5  & 95.8 & 98.8 & \textbf{99.2} \\
		Zipper      & 97.9  & 96.6 & 90.9 & 97.7 \\
		\hline
		Average     & 92.1  & 85.4 & 95.5 & \textbf{97.9} \\
		\hline
	\end{tabular}
\end{table}

\begin{figure}[htbp]
	\centering
	\includegraphics[width=0.46\textwidth]{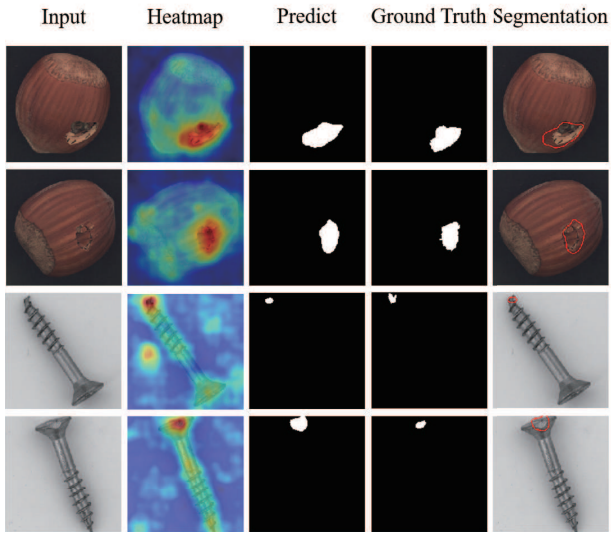}
	\caption{The result on some unaligned dataset by our model.}
	\label{fig:result_show}
\end{figure}

We compared our method against several methods that were introduced over the last several months, as well as longer standing baseline such AE and VAE. Fig. \ref{fig:result_show} shows some visualized results of our proposed method applied on the MVTec AD dataset.

In Tab. \ref{pixel_auc_mvtec}, we compared the pixel level AUROC between our method and other existing state-of-the-art methods for anomaly localization on the MVTec AD dataset. We can observe that 12 out of 15 categories have an accuracy rate greater than 98\%. On average for all classes, our method outperforms all the other methods.
In Tab. \ref{pixel_pro_mvtec}, we compare our method in terms of PRO-score. It is worth mentioning that PRO-score is more meaningful and more judgmental. The metric PRO-score weights ground-truth regions of different sizes equally, which is different from pixel-level AUROC metrics for which one big correctly localization area can make up lots of wrongly localization small area. We can see that our method SOMAD outperforms the second best model PaDiM \cite{Padim} by 1.2\% on average for all classes. Moreover, 13 out of 15 categories have an accuracy rate greater than 90\%.
We also give anomaly scores to entire images to perform anomaly detection at the image level by taking the maximum score of anomaly maps obtained by our model. The Tab. \ref{img_auc} shows that our model SOMAD achieves significantly better results than the existing leveraging pre-trained CNN methods. We notice that our method outperforms the second best model PaDiM \cite{Padim} by 2.4\% on average for all classes of the MVTec AD. Moreover, Our method performs better than other methods in almost all categories.
In particular, some unaligned classes with rotation at different angles, such as Hazelnut, Screw, outperform other methods a large margin (93.3\% vs 100\% for Hazelnut, 84.4\% vs 95.5\% for Screw),  which strongly indicates our approach performs well on unaligned images. 

\subsection{Ablation Study}
\label{ablation study}
We conduct an ablation study on our method in order to find a suitable $k$. We compare the performance using different $k$ in Tab. \ref{ablation_k}. We observe that $k=4$ is the best parameter, the performance descend with increasing $k$. We also explored the effect of SOM map size on the experimental result. As shown in Tab. \ref{tab:ablation_mapsize}, $56 \times 56$ SOM performs best. It is reasonable that the size of SOM is the same as the feature map we used. It makes no sense to continue to increase the size of SOM. Due to space limitations, we only report the average accuracy.

\begin{table}[htbp]
\footnotesize
	\centering
	\caption{ablation experiment results of $k$ (avg. \%).}
	\label{ablation_k}
	\begin{tabular}{c|c|c|c|c|c}
	    \hline
		 $k$    & 1 & 2 & 3 & 4 & 5 \\
		\hline
		PRO-score & 92.4 & 93.2  & 93.2  & \textbf{93.3} & 93.2 \\
		AUROC (image) & 89.3 & 96.9 & \textbf{98.0} & 97.9 & 97.8 \\
		AUROC (pixel) & 97.5 & \textbf{97.8} & \textbf{97.8} & \textbf{97.8} & \textbf{97.8} \\
		\hline
	\end{tabular}
\end{table}
\begin{table}[hbtp]
\footnotesize
	\centering
	\caption{ablation experiment results of SOM size (avg. \%).}
	\label{tab:ablation_mapsize}
	\begin{tabular}{c|c|c|c}
	    \hline
		map size  & $14\times14$ & $28\times28$ & $56\times56$ \\
		\hline
		PRO-score    & 92.1  & 92.8      & \textbf{93.3}      \\
		AUROC (image) & 94.3  & 96.9      & \textbf{97.9}      \\
		AUROC (pixel) & 97.3  & 97.6      & \textbf{97.8}      \\
		\hline
	\end{tabular}
\end{table}

\section{Conclusion}
\label{sec:conclusion}
We have presented a novel method called SOMAD for anomaly detection and localization which is based on self-organizing map. It achieves state-of-the-art performance on MVTec AD datasets. Moreover, our method needs less time consumption and performs better than other methods, especially on non-aligned data, which is more in line with the actual industrial production situation.

{\flushleft\textbf{Acknowledgements}.} This work is funded by National Key Research and Development Project of China under Grant No. 2020AAA0105600 and 2019YFB1312000, National Natural Science Foundation of China under Grant No. 62006183, 62076195 and 62072367, and by China Postdoctoral Science Foundation under Grant No. 2020M683489.

\newpage
\bibliographystyle{IEEEbib}
\bibliography{refs}
\vfill
\pagebreak
%\clearpage
% \newpage

\appendixpage
%\section{Visualization Results}
%\label{appendix}
\begin{figure*}[hb]
	\centering
	\includegraphics[width=13.5cm]{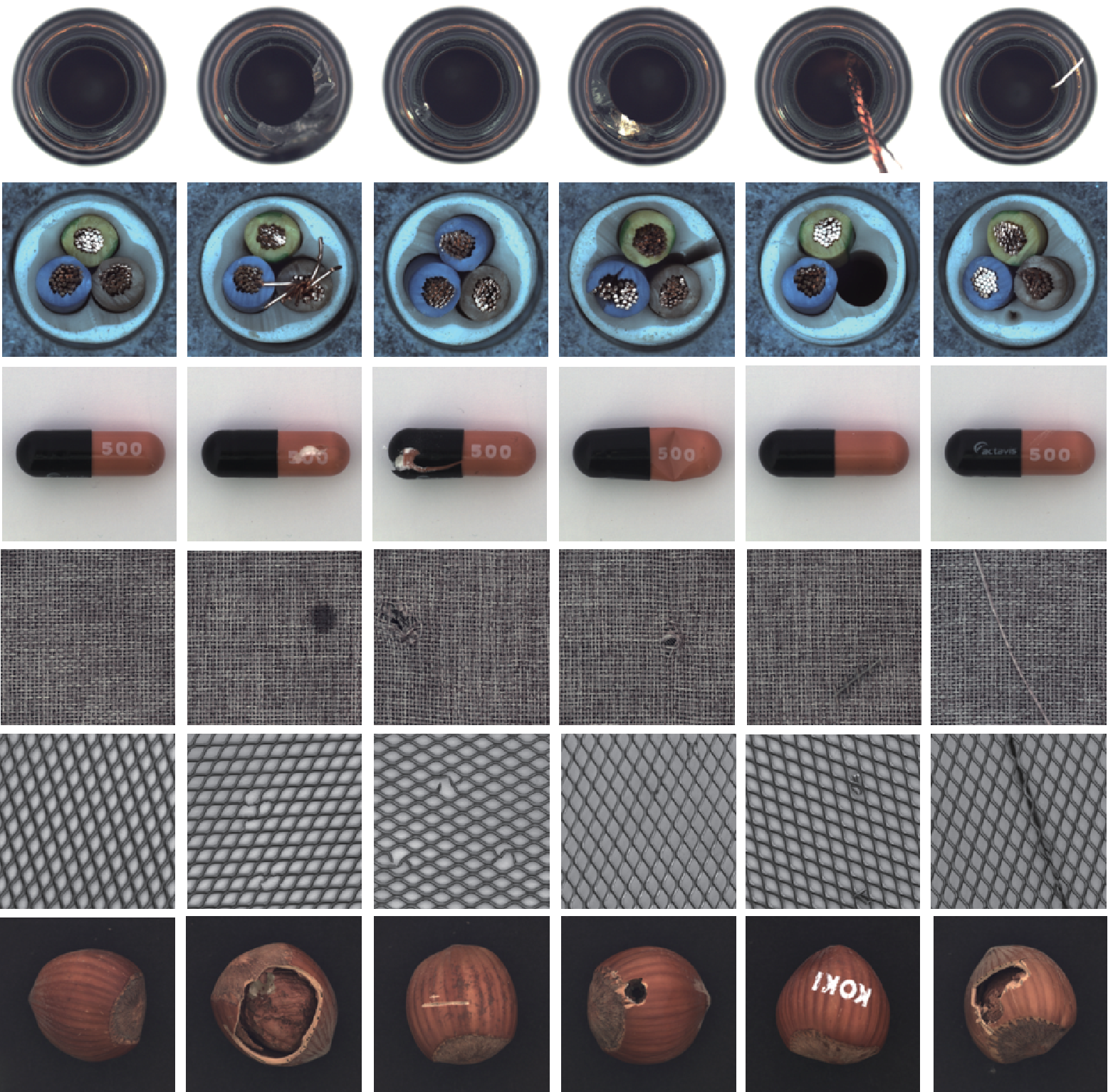}
	\caption{Samples from MVTec AD dataset. The first column is normal, others are defective.}
	\label{fig:mvtec_show1}
\end{figure*}

\begin{figure*}[hb]
	\centering
	\includegraphics[width=13.5cm]{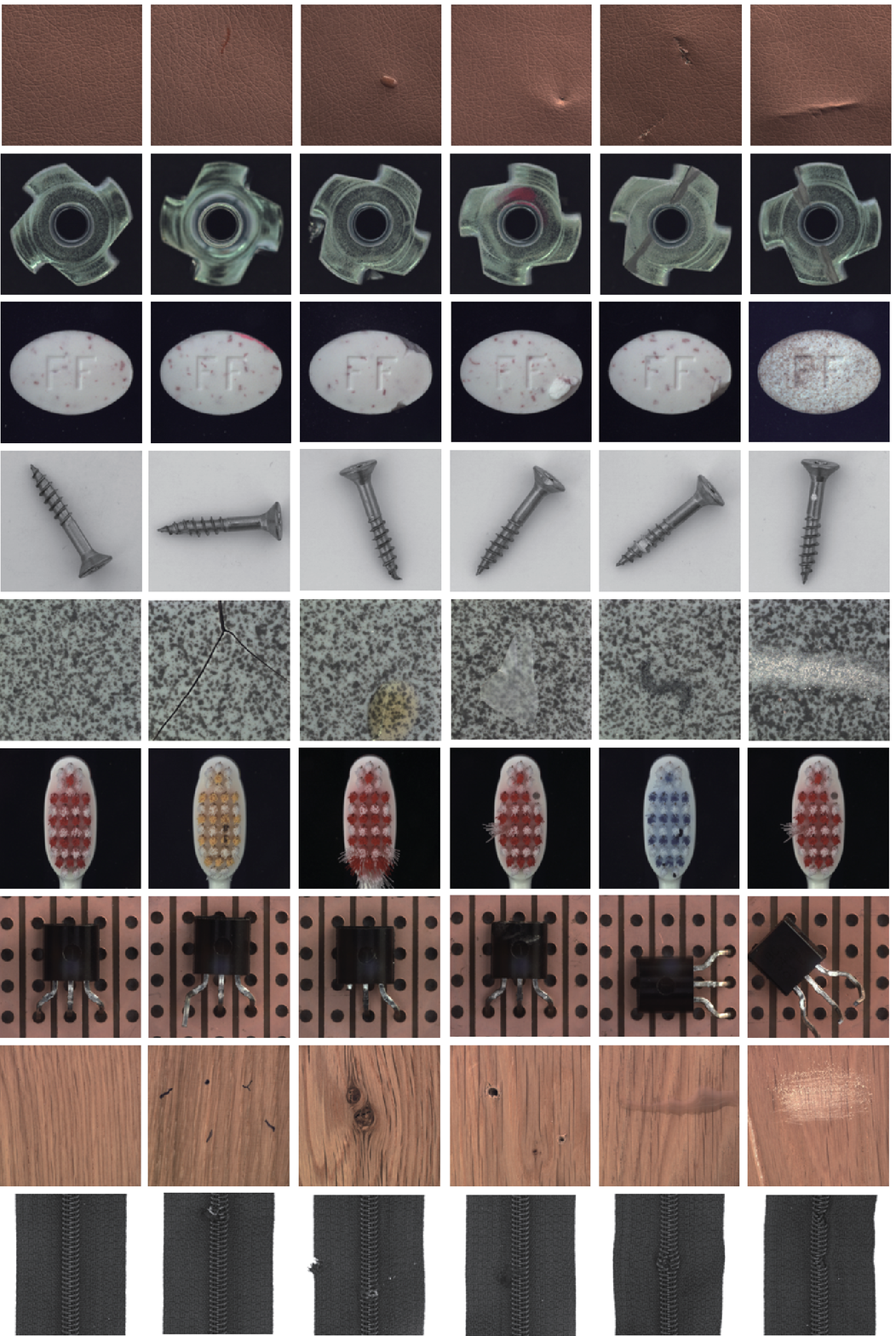}
	\caption{Samples from MVTec AD dataset. The first column is normal, others are defective.}
	\label{fig:mvtec_show2}
\end{figure*}

\begin{figure*}[!b]
	\centering
	\includegraphics[width=13.5cm]{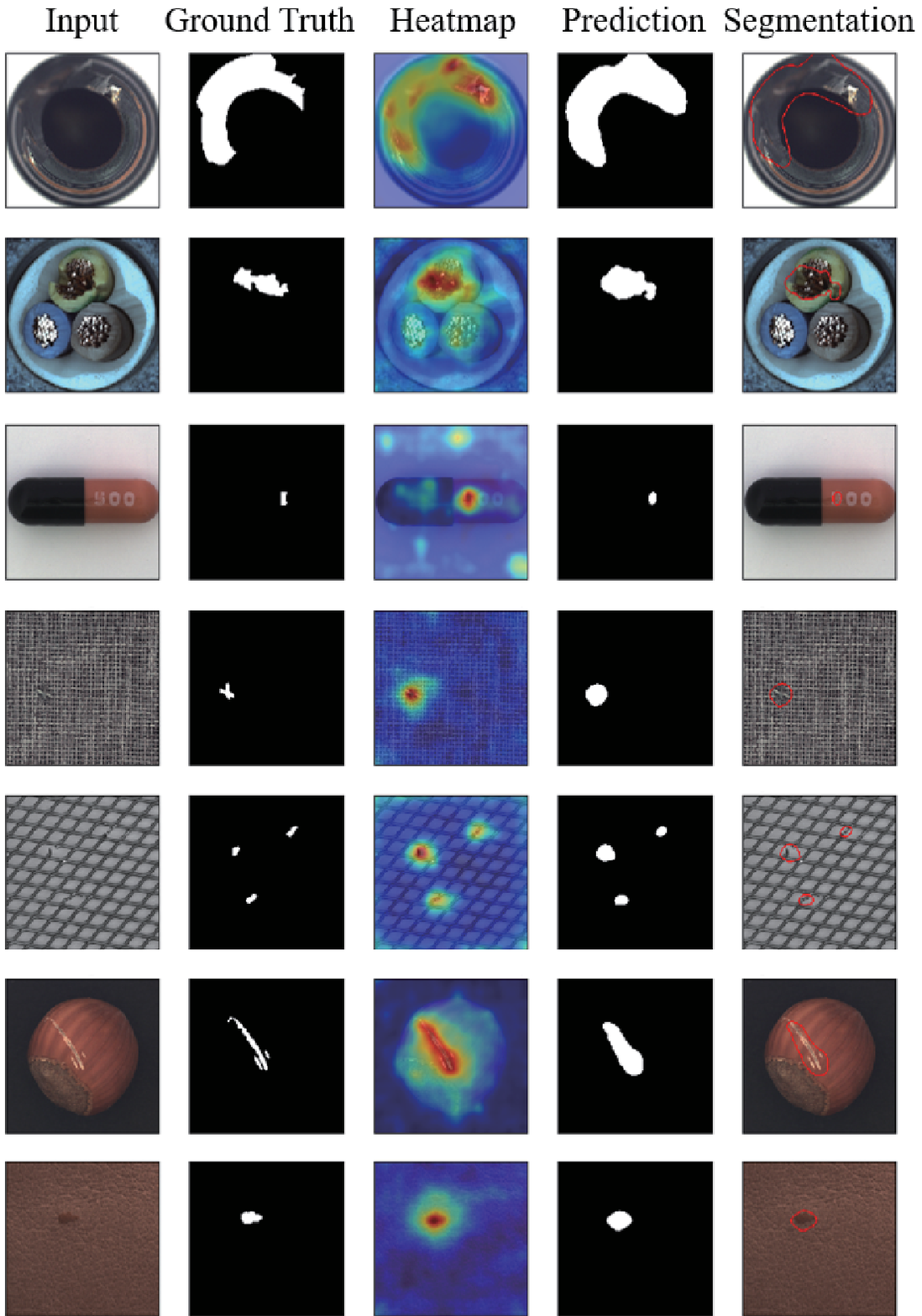}
	\caption{The result on MVTec AD by our model.}
	\label{fig:result_app1}
\end{figure*}

\begin{figure*}[hb]
	\centering
	\includegraphics[width=13.5cm]{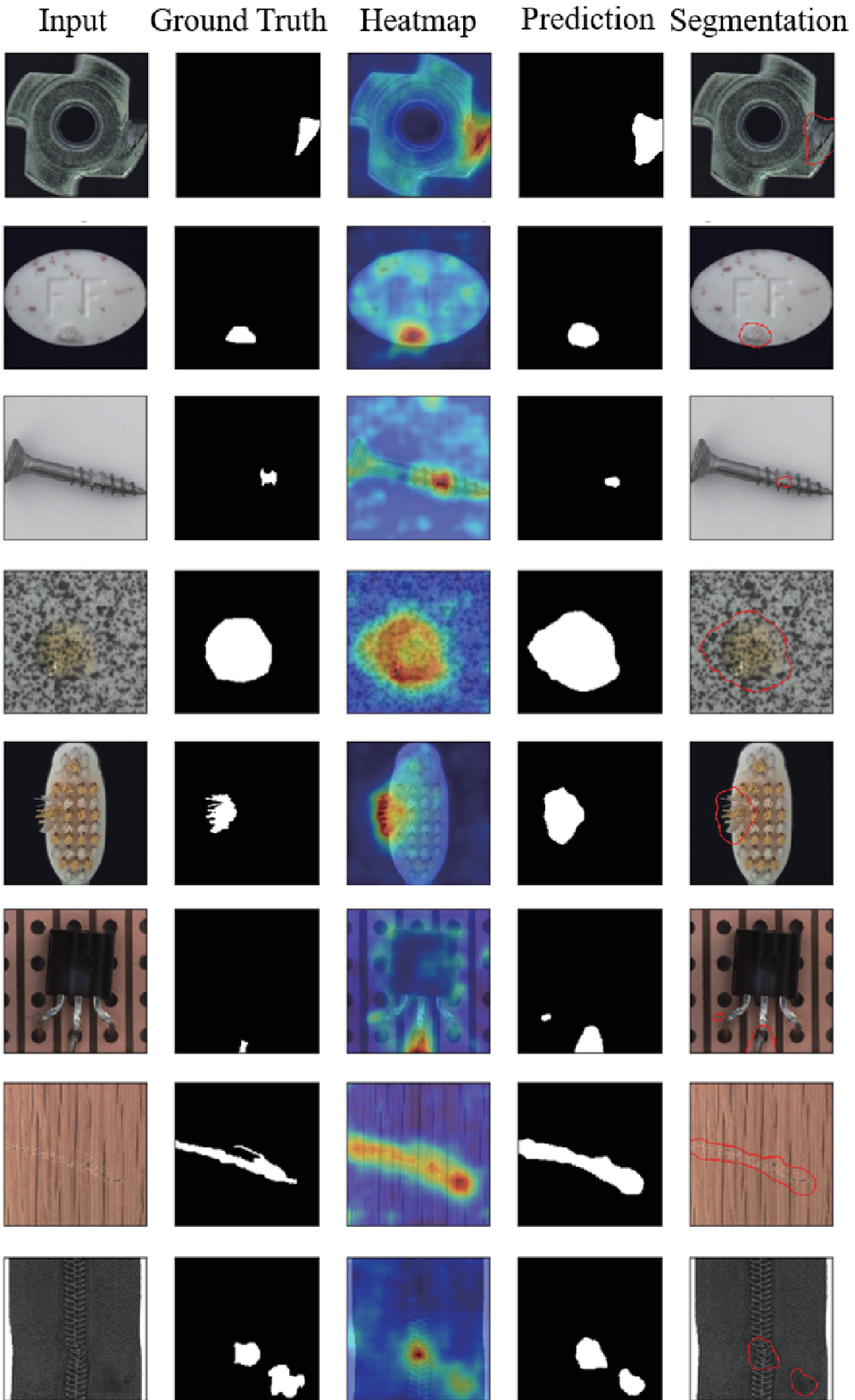}
	\caption{The result on MVTec AD by our model.}
	\label{fig:result_app2}
\end{figure*}
\end{document}